This article has been accepted for inclusion in a future issue of this journal. Content is final as presented, with the exception of pagination.

# Context-Aware Learning for Generative Models

Serafeim Perdikis, *Member, IEEE*, Robert Leeb, *Member, IEEE*,
Ricardo Chavarriaga, *Member, IEEE*, and José del R. Millán, *Fellow, IEEE*

***Abstract*— This work studies the class of algorithms for learning with side-information that emerges by extending generative models with embedded context-related variables. Using finite mixture models (FMMs) as the prototypical Bayesian network, we show that maximum-likelihood estimation (MLE) of parameters through expectation–maximization (EM) improves over the regular unsupervised case and can approach the performances of supervised learning, despite the absence of any explicit ground-truth data labeling. By direct application of the missing information principle (MIP), the algorithms' performances are proven to range between the conventional supervised and unsupervised MLE extremities proportionally to the information content of the contextual assistance provided. The acquired benefits regard higher estimation precision, smaller standard errors, faster convergence rates, and improved classification accuracy or regression fitness shown in various scenarios while also highlighting important properties and differences among the outlined situations. Applicability is showcased with three real-world unsupervised classification scenarios employing Gaussian mixture models. Importantly, we exemplify the natural extension of this methodology to any type of generative model by deriving an equivalent context-aware algorithm for variational autoencoders (VAs), thus broadening the spectrum of applicability to unsupervised deep learning with artificial neural networks. The latter is contrasted with a neural-symbolic algorithm exploiting side information.**

Manuscript received July 15, 2019; revised May 15, 2020; accepted July 12, 2020. This work was supported in part by the European ICT Programme under Project FP7-224631 Tools for Brain-Computer Interaction (TOBI) and in part by the Hasler Foundation, Switzerland. *(Corresponding author: Serafeim Perdikis.)*

Serafeim Perdikis was with the Center for Neuroprosthetics, Chair in Brain-Machine Interface, School of Engineering, Institute of Bioengineering, École Polytechnique Fédérale de Lausanne, CH-1202 Geneva, Switzerland. He is now with the Brain-Computer Interfaces and Neural Engineering Laboratory, School of Computer Science and Electronic Engineering, University of Essex, Colchester CO4 3SQ, U.K. (e-mail: serafeim.perdikis@essex.ac.uk).

Robert Leeb was with the Center for Neuroprosthetics, Chair in Brain-Machine Interface, School of Engineering, Institute of Bioengineering, École Polytechnique Fédérale de Lausanne, CH-1202 Geneva, Switzerland. He is now with Mindmaze SA, CH-1006 Lausanne, Switzerland.

Ricardo Chavarriaga was with the Center for Neuroprosthetics, Chair in Brain-Machine Interface, School of Engineering, Institute of Bioengineering, École Polytechnique Fédérale de Lausanne, CH-1202 Geneva, Switzerland. He is now with the School of Engineering, Institute of Applied Information Technology (InIT), Zurich University of Applied Sciences, CH-8401 Winterthur, Switzerland.

José del R. Millán was with the Center for Neuroprosthetics, Chair in Brain-Machine Interface, School of Engineering, Institute of Bioengineering, École Polytechnique Fédérale de Lausanne, CH-1202 Geneva, Switzerland. He is now with the Department of Electrical and Computer Engineering, The University of Texas at Austin, Austin, TX 78712 USA, and also with the Department of Neurology, The University of Texas at Austin, Austin, TX 78712 USA.

## I. Introduction

THE commonly encountered a situation of missing data labels has raised increasing interest in unsupervised learning approaches for classification. Unsupervised classification can be defined as the task of estimating the parameters of a classification model when the number and type of classes are known, and training data samples are available, but there exist no associated ground-truth data labels whatsoever. The latter characteristic distinguishes this problem from semisupervised learning methods [1], where some labeled instances exist. The absence of any kind of reward signal renders reinforcement learning [2] equally unsuitable. Recent works have showcased that, even in this setting, there exist ways to improve parameter estimation exploiting additional, side information.

Along these lines, this work studies algorithms that can exploit probabilistic contextual information to improve maximum-likelihood estimation (MLE) in generative finite mixture models (FMMs) [3, chap. 9] and variational autoencoders (VAs) [4]. More specifically, we focus on situations where it is possible to extend the probabilistic directed graph of a generative model with contextual random variables $c_i$ whose prior, $p(c_i)$, and/or conditional distributions, $p(z_i|c_i)$ or $p(c_i|z_i)$,[1] are known, thus providing the additional side information. We show that such contextual assistance is able to partially reveal the missing data label information.

As illustrative examples, one can consider an adaptive activity recognition system equipped with online unsupervised learning capabilities to classify a set of activity classes $z$ from kinematic sensor data features $\boldsymbol{x}$. Such a system could benefit from environmental context $c \in \{h(\text{ome}), o(\text{utdoor})\}$, upon which $z$ naturally depends, that is, since the aforementioned distributions [e.g., $p(z = \text{run}|c = o)$ and $p(z = \text{walk}|c = o)$] are indicative of the current activity even in the case of latent context $c$, which is a consequence of the statistical relationships between context and labels, i.e., the facts that running is more likely to occur outdoors while walking indoors. Similarly, unsupervised learning of lung tumor detectors with $z \in \{\text{malignant}, \text{benign}\}$ from X-ray imaging features $\boldsymbol{x}$ could be enhanced by knowledge on the results of a parallel blood test $c$ (observed, but conditionally independent of $\boldsymbol{x}$), where the dependence relationship between $z$ and $c$ is reversed with respect to the previous example.

[1] Variable $z_i$ represents the latent class label of data sample $x_i$.

The main motivation of this study is to show that such algorithms are able to learn "better" than their unsupervised equivalents and close to the supervised ones despite completely discarding any need for ground truth. Second, we wish to explore the information-theoretic principles under which this type of side information yields estimation benefits. The contributions of this article are threefold. First, we draw attention to the fact that simple MLE along with the abovementioned mild assumptions result in improved unsupervised learning, a fact so far neglected in favor of more complex methodologies [5], [6]. Second, for the FMM case, we prove this framework's benefits in various scenarios in terms of parameter estimation precision, standard errors, convergence rates, and classification or regression quality. A comparative analysis of the algorithms in question is also offered. In addition, we demonstrate the applicability of this approach to real-world problems. The third contribution entails the in-depth study of the underlying mechanisms through which these algorithms improve unsupervised estimation, using again the simple FMM case as a vehicle. This includes, on the one hand, the analysis of exemplary likelihood landscapes. On the other hand, we explicitly demonstrate—for the first time—the alleviation of missing label information by side information, through the missing information principle (MIP) [7]. Finally, we demonstrate how the same idea extends to VAs, thus additionally empowering with side-information unsupervised deep learning methods; a topic that despite its recent breakthroughs is still largely dependent on the existence of labeled data sets. Applicability with deep artificial neural networks (ANNs) allows for comparing the proposed approach with recent methods inspired by neural-symbolic integration.

The remainder of this manuscript is organized as follows. Section II discusses the relevant literature and highlights its differences with the present work. Section III presents the examined algorithms, the relevant theory, and the evaluation methodology. Section IV illustrates the results on artificial and real data sets. Finally, Section V discusses this work in light of the results.

## II. Related Work

Most recent work on learning with side information employs undirected graphs and addresses natural language processing (NLP) applications. A great deal of literature addresses various cases of weak supervision where, although some form of data labels is available, it differs from regular supervision. A first case concerns learning from partially or ambiguously labeled data sets, where each data sample is associated with many possible labels only one of which is correct [8], [9]. Second, multilabel, multiannotator (crowd-sourcing) settings where all the labels could be valid, potentially with different or time-varying reliability [10]–[12]. Partial-label problems, where labels are only missing for some of the classes, are studied in [13]. In [14], another partial-label framework is investigated, concerning the case where one knows to which classes a sample does not belong. In addition, multiple-instance or multiview learning methods, where each learning example contains a bag of samples, are proposed in [15]–[18]. In [19], a generic method to handle most of the abovementioned problems is presented. Nguyen *et al.* [20] put forward a framework exploiting additional information in the form of reliability indices of data labels. Similarly, cases with noisy or wrong labels are addressed in [21]–[24]. The setting discussed here differs substantially from all these approaches, as well as from cotraining [25], [26] and all other semisupervised learning methodologies, because in our case the contextual random variables can be virtually anything, including but not restricted to some kind of explicit labeling. Hence, side information on data labels emerges naturally through the dependence relationship between the latent label/class and the contextual variables taking the form of implicit, but not actual "probabilistic labels." Essentially, our framework proposes how "soft" labels can be derived by context without manual effort and explicit labelers.

Another class of related problems regards those where side information is provided in the form of constraints. Most of the early work has focused on known positive and/or negative linkage between pairs or sets of samples [27], [28]. Beyond case-specific methods, there exist frameworks able to cope with context-aware learning irrespective of the form of side information, as elaborated next.

Chang *et al.* [29] proposed constraint-driven learning (CODL), which penalizes constraint violations of a given model by augmenting the objective function with a penalty term. Nevertheless, its formulation requires the existence of some labeled instances for initialization, does not maintain uncertainty during learning, and involves a fairly heuristic optimization algorithm with many hyperparameters. Liang *et al.* [30] put forward a Bayesian approach by modeling side information as the so-called "measurements": noisy expectations of constraint features. The employed objective function is optimized with a complicated variational approximation, which is the method's main disadvantage.

In a series of articles, Mann and McCallum [5] introduced the generalized expectation criteria (GECs), where the additional information comes as linear constraints of a set of feature expectations forming a standalone objective or augmenting the common likelihood objective with an extra term. A special case of GEC had been initially proposed as "expectation regularization" [31]. Several optimization procedures have been presented and tested, including gradient descent [32] and variational approximation [33].

Using the very same modeling of side information, Ganchev *et al.* [6] proposed posterior regularization (PR). In this case, constraints are imposed directly on the posterior distributions of latent models, giving rise to optimization algorithms akin to regular expectation–maximization (EM). PR's conceptual intuitiveness has contributed to its recent popularity [34]–[36]. Ghosh *et al.* [37] independently proposed a PR formulation specific to FMMs and constraints in the form of *a priori* knowledge of mixing proportions, deriving a variant of the "scaled"-PR algorithm for this particular problem [6, Appendix A]. Despite sharing the same model, this work exploits a less generic type of side information and involves complex formulations.

In a brilliant analysis [6, Sec. 4], it is shown that under certain approximations, all four generic frameworks are equivalent. Compared with these approaches, it can be said that the algorithms examined here trade off generality in favor of simplicity and intuitiveness. This claim is substantiated in Appendix A (see the Supplementary Material), where the PR equivalents of our algorithms are discussed.

The idea of augmenting a given model to include context, the cornerstone of our work, can be traced back to the "hierarchical shrinkage" method [38]. Probabilistic context modeling identical to ours is proposed in [39]. However, in this case, the authors focus on classification improvements rather than the estimation properties of the algorithm.

Some of the aforementioned studies propose algorithms with identical formulations to those proposed in the present work. Specifically, Bouveyron and Girard [21] and Côme *et al.* [40] produced the formulation of what we call here the WCA algorithm, in the context of learning with noisy labels and through Dempster–Shafer basic belief assignments, respectively. On the other hand, Ambroise and Govaert [14] and Szczurek *et al.* [22] (who also compared with the work of Côme *et al.*) arrived at the formulation of our CA algorithm assuming, again, the existence of "soft" supervision. Our work is, first, more general than those since we exhaustively compare all algorithmic possibilities. Most importantly, as already mentioned, our derivations do not take the existence of "uncertain" labels for granted and discard the need for any kind of ground truth. In addition, the scope of this article is the only one strongly focused on the information-theoretic effects of learning with side information. Finally, we show extensibility to different generative models, including deep neural nets.

Given the recent advent of deep learning and the central role of ANNs therein, methods born in the framework of neural-symbolic integration [41] (where data-driven inference is combined with independent, "symbolic" knowledge—e.g., logic rules) have been considered, where symbolic knowledge plays the role of side information. In the two most relevant methods, Hu *et al.* [42] used knowledge distillation (KD) to harness supervised learning with logical constraints; however, this method still mainly relies on the availability of labels, with rules only refining the learning process. The idea of augmenting deep generative models with side information can be found in [43], but the authors do not go beyond the specific setting they discuss (oracle-provided similarity constraint triplets) and focus on explainability rather than performance.

## III. Methods

### A. Context-Aware Learning Algorithms for FMMs

The proposed idea is best explained in the framework of FMMs, the simplest and most basic Bayesian network. In order to gain a solid understanding, the reader should recall [3, Ch. 9.2] that an FMM is represented by the directed graph shown in Fig. 1(a) and (b) (enclosed in a dashed box), where $\boldsymbol{x_i} \in X$ is the observed independent and identically distributed (i.i.d.) data samples of a data set $X$ with cardinality $N$ ($i \in$ [1,N]), $\boldsymbol{z_i} \in Z$ is the latent data representing the mixture/class generating sample $\boldsymbol{x_i}$ having a 1-of-$M$ representation, so that $z_{ij} \in \{0, 1\}$ and $\sum_j z_{ij} = 1$, and $M$ is the number of mixtures/classes. The distribution of observed data $\boldsymbol{x}$ is then $p(\boldsymbol{x}) = \sum_z p(\boldsymbol{x}, z) = \sum_z p(z) p(\boldsymbol{x}|z) = \sum_{j=1}^{M} \pi_j f_j(\boldsymbol{x}, \boldsymbol{\theta_j})$, where $\pi_j = p(z_j = 1)$ are the mixture coefficients with $\sum_{j=1}^{M} \pi_j = 1$ and $f(\boldsymbol{x}, \boldsymbol{\theta}') = p(\boldsymbol{x}|z, \boldsymbol{\theta}')$ with $f$ belonging to some identifiable family with parameters $\boldsymbol{\theta}'$. ML estimation consists in maximizing the logarithm of the incomplete data, marginal likelihood $\log L(\boldsymbol{\theta}|X) = \log(\prod_{i=1}^{N} p(\boldsymbol{x_i}))$ over $\boldsymbol{\theta}$. In supervised estimation, $\boldsymbol{z_i}$ are the observed labels $\boldsymbol{y_i}$, yielding analytic solutions. Conversely, for latent $\boldsymbol{z_i}$, one relies on the iterative EM-MLE, where, first, the expectation (under posteriors $p(\boldsymbol{z}|\boldsymbol{x}, \boldsymbol{\theta})$) of the complete-data log-likelihood $\log L_c(\boldsymbol{\theta}|X, Z)$ is formed (E-step)

$$Q\left(\boldsymbol{\theta}, \hat{\boldsymbol{\theta}}^k\right) = \mathbb{E}_{\hat{\boldsymbol{\theta}}^k}\{log L_c(\boldsymbol{\theta}|X, Z)\} = \sum_{i,j}^{N,M} \mathbb{E}_{\hat{\boldsymbol{\theta}}^k}\{z_{ij}\} \log \pi_j + \sum_{i,j}^{N,M} \mathbb{E}_{\hat{\boldsymbol{\theta}}^k}\{z_{ij}\} \log\left(f_j\left(\boldsymbol{x_i}, \boldsymbol{\theta_j}\right)\right) \tag{1}$$

where $\boldsymbol{\theta} = \{\pi_j, \boldsymbol{\theta_j}\}, \forall j$ are the overall estimated parameters and $\hat{\boldsymbol{\theta}}^{\boldsymbol{k}}$ are the $k$th estimate. Then, $Q(\boldsymbol{\theta}, \hat{\boldsymbol{\theta}}^{\boldsymbol{k}})$ can be analytically maximized (M-step): $\hat{\boldsymbol{\theta}}^{\boldsymbol{k+1}} = \underset{\boldsymbol{\theta}}{\operatorname{argmax}}\{Q(\boldsymbol{\theta}, \hat{\boldsymbol{\theta}}^{\boldsymbol{k}})\}$.

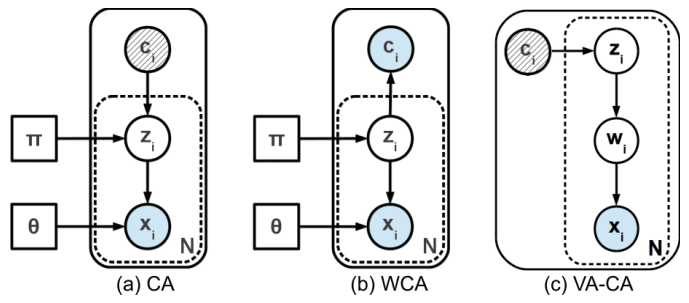

Fig. 1. Graphical representations of augmented (solid boxes) and regular (dashed boxes) mixture models for a set of $N$ i.i.d. data samples. Random variables depicted in circles, transparent for latent variables, shaded for observed variables and stripping for variables that can be observed or latent on occasion. Model parameters are illustrated with squares. $\boldsymbol{x_i}$ are the observed data samples, $z_i$ are the latent class labels, and $c_i$ are the contextual variables. Model (a) gives rise to CA type of estimation and (b) to WCA. Model (c) depicts the graph of a VA-based mixture for the equivalent of CA estimation.

This conventional unsupervised EM-MLE algorithm (termed hereafter US) is known to get stack in local maxima (thus being sensitive to the initialization $\hat{\boldsymbol{\theta}}^{\boldsymbol{0}}$) and exhibits compromised estimation precision compared with supervised estimation (termed S). Furthermore, it is inferior to S in terms of standard errors and convergence rate (since it is iterative). It is clear that these limitations should be related to the missing label information. Both methods share the same objective of (1), only differing in the replacement of labels $\boldsymbol{y_i}$ (S) by posteriors $\mathbb{E}_{\hat{\boldsymbol{\theta}}^k}\{z_{ij}\} = p(\boldsymbol{z_i} = j|\boldsymbol{x_i}, \hat{\boldsymbol{\theta}^k})$ (US). Hence, it is reasonable to assume that boosting the information content (entropy) of distributions $\mathbb{E}_{\hat{\boldsymbol{\theta}}^k}\{z_{ij}\}$ toward the labels $\boldsymbol{y_i}$ should raise US performances closer to those of S.

The idea put forward in this article is to achieve this goal by directly embedding probabilistic side information into a generative model. More specifically, it suffices that contextual information can be modeled by (in general, latent) random

TABLE I

ALGORITHMS UNDER STUDY. US: REGULAR UNSUPERVISED EM-LEARNING. CA: CONTEXT-AWARE EM-LEARNING. WCA: WEIGHTED CONTEXT-AWARE EM-LEARNING. DCA: DIRECT CONTEXT-AWARE LEARNING. S: REGULAR SUPERVISED LEARNING

| | $\boldsymbol{p_i}$ | E–step $\mathbb{E}_{\hat{\boldsymbol{\theta}}}\{z_{ij}\} = \dots$ | $logL = \sum_{i=1}^{N} \log(\sum_{j=1}^{M}(\dots))$ |
|---|---|---|---|
| *US* | None | $\frac{\pi_j f_j(\boldsymbol{x_i}\|\hat{\boldsymbol{\theta}}_{\boldsymbol{j}})}{\sum_{m=1}^{M} \pi_m f_m(\boldsymbol{x_i}\|\hat{\boldsymbol{\theta}}_{\boldsymbol{m}})}$ | $\pi_j f_j(\boldsymbol{x_i}\|\hat{\boldsymbol{\theta}})$ |
| *CA* | $\begin{cases} \sum_{c_i} p(c_i)p(\boldsymbol{z_i}\|c_i), & \text{latent C} \\ p(\boldsymbol{z_i}\|c_i), & \text{observed C} \end{cases}$ | $\frac{p_{ij} f_j(\boldsymbol{x_i}\|\hat{\boldsymbol{\theta}}_{\boldsymbol{j}})}{\sum_{m=1}^{M} p_{im} f_m(\boldsymbol{x_i}\|\hat{\boldsymbol{\theta}}_{\boldsymbol{m}})}$ | $p_{ij} f_j(\boldsymbol{x_i}\|\hat{\boldsymbol{\theta}})$ |
| *WCA* | $\frac{p(c_i\|\boldsymbol{z_i})}{p(c_i)}$ | $\frac{p_{ij}\pi_j f_j(\boldsymbol{x_i}\|\hat{\boldsymbol{\theta}}_{\boldsymbol{j}})}{\sum_{m=1}^{M} p_{im}\pi_m f_m(\boldsymbol{x_i}\|\hat{\boldsymbol{\theta}}_{\boldsymbol{m}})}$ | $p_{ij}\pi_j f_j(\boldsymbol{x_i}\|\hat{\boldsymbol{\theta}})$ |
| *DCA* | Custom | $p_{ij}$ | $\pi_j f_j(\boldsymbol{x_i}\|\hat{\boldsymbol{\theta}})$ |
| *S* | $y_i$ | $\begin{cases} 1 & , y_i = j \\ 0 & , y_i \neq j \end{cases}$ | $\pi_j f_j(\boldsymbol{x_i}\|\hat{\boldsymbol{\theta}})$ |

variables $c_i$,[2] which can be assumed to have a dependence relationship with the latent nodes $\boldsymbol{z_i}$ (augmenting the underlying model, as shown in Fig. 1 for the case of FMMs) and whose distributions $p(c_i)$ and/or $p(\boldsymbol{z_i}|c_i)$ and $p(c_i|\boldsymbol{z_i})$ are known. Given these prerequisites, deriving context-aware algorithms results from the straightforward application of MLE on the augmented models, which results in the regular EM algorithm for FMMs. Analytical derivations can be found in Appendix B (see the Supplementary Material).

It is critical to discuss what these assumptions imply for the applicability of the proposed approach. The first prerequisite is a mere modeling choice and hardly restrictive since all natural quantities can be modeled as random variables. The second assumption forms the basis of our framework. It advocates for a paradigm shift where one needs not solely rely on the possibility to collect usual "data and labels," but can instead identify contextual sources of information that may partially reveal the missing data labels. Of course, this might not always be possible. The third assumption can also be limiting since, even after identifying potentially useful types of context, the distributions $p(c)$, $p(z|c)$, $p(c|z)$ could still be unknown, difficult to preestimate, or rather uninformative. For instance, in the medical informatics example used in Section I, medical tests additional to X-ray imaging for lung tumor detection might be as hard, expensive, or dangerous to collect as the biopsy that would reveal the actual ground-truth labels (malignant or benign). In addition to this, this last prerequisite can only be satisfied for fully defined $z$ (i.e., known number of mixtures $M$), which limits the scope to unsupervised classification. Hence, with respect to the last two assumptions, there can be no guarantees of existence or benefits or even a way to analytically quantify the likelihood of those in general applications. Nevertheless, in the era of information explosion and the emergence of the Internet of Things, we believe that these assumptions can be already satisfied more often than not, with the situation improving in the foreseeable future.

[2]Without loss of generality, the contextual random variables will be assumed hereafter to be univariate and discrete.

The two possible types of dependence between $c_i$ and $\boldsymbol{z_i}$ give rise to two different augmented models [see Fig. 1(a) and (b)] and, therefore, two corresponding EM-MLE algorithms termed CA and WCA, respectively. We are also considering a third, heuristic algorithm termed DCA (direct context aware), where the posterior distribution of latent labels is defined by arbitrary probabilistic labels and the evidence $X$ is ignored. Table I summarizes the naming convention, probabilistic labels $\boldsymbol{p_i}$, E-step, and log$L$ formulation of each algorithm.

From a factor graph perspective [3, Ch. 8.4.3], the basic premise of CA and WCA is the provision of additional (compared with US) information through the messages passed to the latent nodes $z_i$. As evident in the E-steps of Table I, US only benefits from evidence $X$, while belief propagation with CA and WCA should be richer due to the additional contextual variables $C$. Of note, in the original S/US estimation problem, $\boldsymbol{\theta}$ is not cumbered with additional parameters related to the contextual variables, despite the model augmentation, due to the assumption of known priors and conditionals. It follows that the graphical representation of contextual assistance can be more complex than a single variable $c$, as long as the conditions of no additional parameters and seamless message parsing are satisfied. Essentially, the need for ground truth is replaced by a lesser requirement for knowledge of the aforementioned distributions. Those can be learned prior to the deployment of the algorithms or even be publicly available (e.g., language models).

As already illustrated (see Table I), the context-related terms of each algorithm can be isolated to implicitly define samplewise probabilistic "labels" $p_{ij}$ with $\sum_{j=1}^{M} p_{ij} = 1$ (i.e., each $\boldsymbol{p_i}$ is a discrete probability distribution over the latent variable $\boldsymbol{z_i}$). The entropy of these labels represents a measure of the contextual information content individually for each sample $\boldsymbol{x_i}$ and, by averaging, for the overall estimation problem. Our work is the only one, besides [40], highlighting the importance of side-information measurability for the prediction of estimation benefits, our primary axis of investigation.

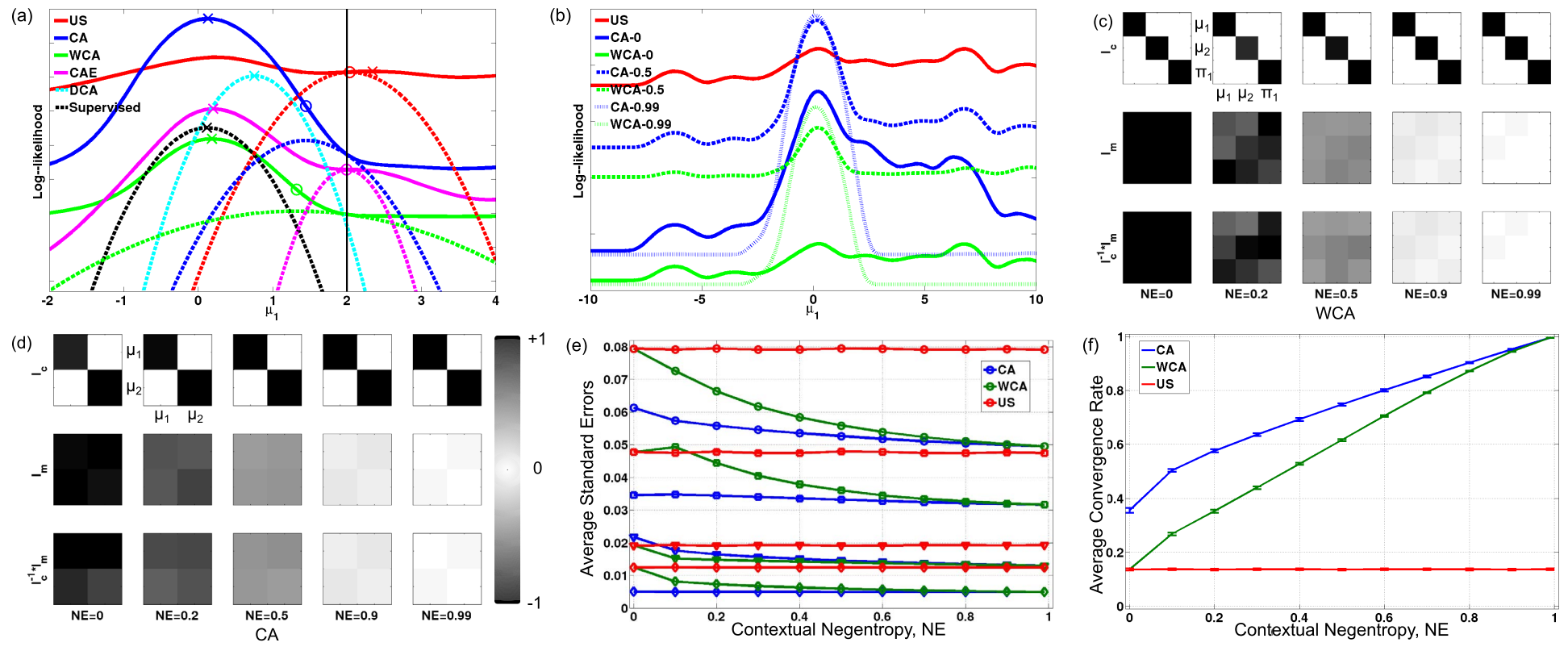

Fig. 2. (a) $\log L(\mu_1)$ (solid lines) and $Q(\mu_1, \hat{\mu_1}^0) + H(\hat{\mu_1}^0, \hat{\mu_1}^0)$ (at first EM iteration, dashed lines) for various estimation algorithms as color-coded in the legend, in a mixture of two univariate Gaussians model. CAE refers to algorithm CA with ignorant context and CA, WCA, and DCA are shown for contextual negentropy NE = 0.7. ML estimates after convergence shown with "x" and estimates after the first iteration in "o." (b) Incomplete-data log-likelihoods $\log L(\mu_1)$ in the same problem as in (a) for US (red solid line) and various contextual negentropy levels as shown in the legend for CA (blue line) and WCA (green line). Information matrices $I_c$ and $I_m$ and rate matrix $J = I_c^{-1} * I_m$ with increasing contextual negentropy NE in a mixture of two univariate Gaussians' estimation problem (see text) for (c) WCA and (d) CA. All matrix values normalized to $[-1, 1]$ by dividing with the respective absolute value of the US algorithm. (e) Average predicted standard errors of $\hat{\pi_1}$ (◇), $\hat{\mu_1}$ (▽), and $\hat{\mu_2}$ (□) and their sum (○) with increasing contextual negentropy NE in 100 repetitions of a mixture of two univariate Gaussians' estimation problem. Different algorithms color-coded in the legend. (f) Average predicted convergence rates and their standard deviations for the three algorithms (color-coded in the legend) in the same estimation problem.

### *B. Information Matrices and MIP*

In order to shed light on the fundamental issue of alleviating the missing label information through the provision of side information, one can rely on the Fisher Information [44], the most formal way of measuring the amount of information involved in the estimation of the unknown parameters $\boldsymbol{\theta}$. Therefore, we study approximations of the (expected) Fisher information matrix $I(\boldsymbol{\theta})$ through its sampled-based version, the observed information matrix $I(\boldsymbol{\theta}|X)$. The latter measures the amount of information a sample $X$ carries on the estimated parameters $\boldsymbol{\theta}$, where $I(\boldsymbol{\theta}) = \mathbb{E}_{\hat{\boldsymbol{\theta}}}[I(\boldsymbol{\theta}|X)]$ and $I(\boldsymbol{\theta}|X) = -\partial log L(\boldsymbol{\theta})/\partial\boldsymbol{\theta}\partial\boldsymbol{\theta}^T|_{\boldsymbol{\theta}=\hat{\boldsymbol{\theta}}_{\text{ML}}}$, the negative of the Hessian of the log-likelihood objective function evaluated at the ML estimate.

In [7], it is proved that the observed information for missing-data problems can be computed as the difference $I(\boldsymbol{\theta}|X) = I_c(\boldsymbol{\theta}|X) - I_m(\boldsymbol{\theta}|X)$. The first term is an estimate of the available information if there were no missing data. The second term, called the missing information matrix, represents the information lost due to missing data. This relation has been called the MIP. Both these matrices can be computed through complete-data quantities (so that their calculation is tractable), as $I_c(\boldsymbol{\theta}|X) = \mathbb{E}_{\hat{\boldsymbol{\theta}}}\{-\partial^2 \, log L_c/\partial\boldsymbol{\theta}\partial\boldsymbol{\theta}^T\}|_{\boldsymbol{\theta}=\hat{\boldsymbol{\theta}}_{\text{ML}}}$ and $I_m(\boldsymbol{\theta}|X) = \text{cov}_{\hat{\boldsymbol{\theta}}}\{\boldsymbol{S_c}(X|\boldsymbol{\theta})\boldsymbol{S_c}(X|\boldsymbol{\theta})^T\}|_{\boldsymbol{\theta}=\hat{\boldsymbol{\theta}}_{\text{ML}}}$, where $\boldsymbol{S_c}(X|\boldsymbol{\theta})$ is the score (gradient vector) of the complete-data log-likelihood.

The Fisher information also allows for the computation of the variance–covariance matrix of the MLE, as $C = I^{-1}(\boldsymbol{\theta}|X)$ and, hence, the standard errors of parameter estimation as $\text{SE}_i = 2(I_{i,i}^{-1}(\boldsymbol{\theta}|X))^{1/2}$ for the $i$th parameter in vector $\boldsymbol{\theta}$, without resorting to repeated sampling. The same is true for the algorithms' convergence rate, since when EM converges to a local maximum, it has been shown [45] that the convergence rate $r = \lim_{k\to\infty} \|\hat{\boldsymbol{\theta}}^{k+1} - \hat{\boldsymbol{\theta}}^k / \hat{\boldsymbol{\theta}}^k - \hat{\boldsymbol{\theta}}^{k-1}\|$ is linear and coincides with the spectral radius ($\lambda_{\max}$, where $\lambda_i \in [0, 1), \forall i$, the eigenvalues) of the "rate" matrix $J$, defined as $J(\boldsymbol{\theta}) = I_c^{-1}(X|\boldsymbol{\theta})I_m(X|\boldsymbol{\theta})$. The latter expresses the total fraction of missing information [46]. In Section IV, we use the definition $r' = 1 - r = 1 - \lambda_{\max}$, which complies with the intuition that 0 corresponds to nonconverging and 1 to immediately converging algorithms.

### *C. Likelihood Landscapes*

Fig. 2(a) shows the incomplete-data log-likelihoods $\log L(\mu_1)$ and the intermediate objectives maximized at the first M-step (complete-data log-likelihood expectations plus the latent data entropy) $Q(\mu_1, \hat{\mu_1}^0) + H(\hat{\mu_1}^0, \hat{\mu_1}^0)$ (see Appendix B) for a mixture of two univariate Gaussians.[3] The estimation precision benefits of context-aware algorithms (in case of "correct" context as in the example) come as a result of a favorably modified $\log L(\boldsymbol{\theta})$ (compared with that of US). These log-likelihoods (here with NE = 0.7) tend to have a larger local maximum close to the supervised estimate (black "x") and, hence, also closer to the true parameter value. This maximum will thus also tend to be the global maximum. Even more, the other local maxima are suppressed, reducing the sensitivity to initialization (where US is known to perform poorly). In Fig. 2(a), all context-aware algorithms "escape" (in contrast to US) the local maximum of US on the right side of $\hat{\mu_1}^0$ and converge to the first local maximum on the left side ("x" of the respective color). The latter is much closer to the true parameter value, yielding higher estimation precision.

[3]Only $\mu_1$ is estimated starting from $\hat{\mu_1}^0 = 2$. Parameters $\pi_1 = 0.1, \mu_2 = 1, s_1 = 0.5, s_2 = 3$ are fixed. Estimation is based on $N = 100$ randomly generated samples with $\mu_1 = 0$. This setting is selected among those leading to multiple local maxima so that the effects of the proposed algorithms can be clearly illustrated.

The magnitude of these effects increases with the increase in contextual negentropy NE, as shown in Fig. 2(b). As NE $\rightarrow$ 1 (dotted lines), the problem reduces to supervised learning. Furthermore, in the case of ignorant context (NE = 0), the WCA objective (solid green line) reduces to a translated version of that of US (red line), whereas CA (blue line) already "boosts" the favorable maximum. Hence, for "ignorant" context, WCA is identical to US, whereas CA can already yield improvements in precision.

The improvement of convergence rate and its dependence on contextual negentropy is also implied in the example since the estimates $\hat{\mu}_1^1$ (first iteration) for CA (blue "o") and WCA (green "o") are much closer to their final MLE than for US (red line) or CAE (CA with ignorant context, magenta). This fact is further substantiated in Section III-D. DCA, as elaborated later, underperforms in comparison with the other context-aware algorithms, suggesting that discarding data-dependent evidence completely is suboptimal.

### D. Standard Errors and Convergence Rate Through the MIP

The application of the MIP on algorithms CA, WCA, and US is demonstrated in a (randomly selected) mixture of two univariate Gaussians' problem.[4] For each (increasing) NE value, we perform 100 repetitions randomly generating $N = 10^4$ samples from the above distribution and estimate the algorithms' average (across repetitions) standard errors and convergence rates.

For one repetition of the aforementioned problem, Fig. 2(c) (for WCA) and 2(d) (for CA) shows that matrices $I_c$ (first row) remain unaffected by increasing NE since the complete-data information $I_c = I^S$ should be independent of any side information. On the contrary, the magnitudes of the elements of the missing information matrix $I_m$ (second row), which for NE = 0 obtains its maximum $I_m^{\text{MAX}} = I_m^{US}$ (when no additional information on missing labels exists), are reduced with increasing NE, to eventually vanish into the zero matrix as NE $\rightarrow 1$ ($I_m^{\text{MIN}} = I_m^S = 0$ since data labels are fully revealed by context). Consequently, the fractions of missing information (rate matrix $J = I_c^{-1} * I_m$, third row) also vanish, along with its spectral radius. Hence, the missing label information in context-aware EM learning is shown to be eliminated proportionally to the information content of the provided side information.

The finally available information in these algorithms, as encoded in the information matrices $I = I_c - I_m$ (MIP definition), will consequently be bounded. Above, by $I^{\text{MAX}} = I_c^{\text{MAX}} - I_m^{\text{MIN}} = I_c^S - 0 = I^S$ (identical to the supervised MLE, S) and below by, $I^{\text{MIN}} = I_c^{\text{MIN}} - I_m^{\text{MAX}} = I_c^S - I_m^{\text{US}} = I^{US}$ (identical to the unsupervised MLE, US). Expressions $I_c^{\text{MAX}} = I_c^{\text{MIN}} = I^S$ and $I_m^{\text{MAX}} = I_m^{US}$ and $I_m^{\text{MIN}} = I_m^S = 0$ hold by definitions. These exemplary results are substantiated and shown to generalize to arbitrary FMM problems by Propositions 1–3 of Appendix D (see the Supplementary Material).

Fig. 2(e) shows the average predicted standard errors of estimates $\hat{\pi}_1$ ($\Diamond$), $\hat{\mu}_1$ ($\triangledown$), and $\hat{\mu}_2$ ($\Box$) as well as their sum ($\circ$) with increasing contextual negentropy NE for US (red line), CA (blue line), and WCA (green line). Indeed, it is illustrated that standard errors of all parameters (see below for the explanation of exceptions), as well as their sum, decrease with increase in NE for both CA and WCA, converging toward the corresponding standard errors of S at NE = 1. Similarly, Fig. 2(f) shows that the convergence rate $r'$ of CA (blue line) and WCA (green line) is improved with increase in NE to reach single-iteration convergence ($r' = 1$, like S) in the case of "perfect" context (NE = 1). Thus, as expected (Section III-B), stronger side information improves the performances on these two metrics.

Interestingly, the standard errors and convergence rates of CA are superior to their WCA equivalents for the whole NE spectrum, until both algorithms converge to their $S$ limit at NE $\rightarrow$ 1. CA thus yields a better lower bound than the regular US performances, exhibited by WCA. This is due to the structure of the information matrices for CA which, compared with those of US/WCA, are reduced by removing the elements associated with the mixing coefficients $\pi_j, \forall j$ [as in Fig. 2(d)]. Essentially, in CA, the mixing coefficient distribution $\boldsymbol{\pi}$ is independent of $X$ and fully determined by $C$. The mixing coefficients $\pi_j$ are computed here for CA as $\pi_j = 1/N \sum_i p'_{ij}$, where $p'_{ij} = 1$ if $j = \text{argmax}_k\{p_{ik}\}$ and 0 otherwise, an unbiased estimator exhibiting the same standard errors for $\pi_j$ as S ("correct" context). The fraction of missing information is always reduced for CA, due to the absence of uncertainty regarding $\boldsymbol{\pi}$.[5] The lower bound of CA still corresponds to that of US, but the one with fixed $\boldsymbol{\pi}$.

It is also interesting that, exceptionally, the standard errors of $\mu_1$ [see Fig. 2(e)] are superior (smaller) for WCA rather than CA. Such exceptions can occur because the missing information is not distributed uniformly across the estimated parameters or identically among the different algorithms. However, the sum/average of standard errors only depends on the overall fraction of missing information ($\lambda_{\max}$ of $J$), shown to always reduce with an increase in NE (see Proposition 2, Appendix D) and be smaller for CA compared with WCA. Furthermore, since the rates of convergence coincide to a global rate only depending on $\lambda_{\max}$ (except for rare cases [48]), the rate of convergence of CA is also guaranteed to be greater/equal to that of US and WCA (for a given NE).

### E. Context-Aware Learning Algorithms for VAs

Formulating the proposed algorithms for other Bayesian networks follows the exact same steps. For example, context-aware learning of Hidden Markov Model (HMM) parameters would exploit contextual variables codependent to the hidden state (with known statistics) and result in a version of the Baum–Welch algorithm with modified E-step to include side information and identical M-step. However, in view of classification tasks, it would be ideal to be able to apply the proposed approach in the context of deep learning with ANNs, which have shown to excel in this task when big data are present due to far superior expressiveness than FMMs or other probabilistic generative models. This possibility would also alleviate a major weakness of this literature, namely, the heavy reliance on supervised learning which, alongside the need for (labeled) big data, limits their wider deployment.

[4] $\pi_1 = 0.6$, $\mu_1 = 0$, $\mu_2 = 1$, $s_1 = 1$, and $s_2 = 2$. Parameters $s_1$ and $s_2$ are fixed. $\pi_1$, $\mu_1$, and $\mu_2$ are estimated from initial $\hat{\pi}_1{}^0 = 0.5$, $\hat{\mu}_1{}^0 = 0.49$, and $\hat{\mu}_2{}^0 = 0.51$.

[5] In general, it can be shown by applying [47, Th. 4] that removing any parameter from the estimation problem can only reduce the spectral radius of the rate matrix $J$.

At first, this endeavor may seem odd, with several reasons ostensibly preventing the derivation of equivalent context-aware algorithms in this framework. First, contextual random variables cannot be intuitively embedded into nonprobabilistic models, such as ANNs. Most importantly, these are discriminative models conventionally requiring loss functions other than MLE, which demand explicit labels ("targets"). Furthermore, by avoiding to model the data and joint distributions, they yield no regular unsupervised learning algorithm to build upon. Yet, by imposing softmax output layers, ANNs can, in fact, represent conditional distributions and can be trained with MLE. In addition to this, generative deep neural networks admitting probabilistic graphical representations have been recently formulated, the most popular of which is termed VA [4].

Hence, the basic idea is to derive a mixture model with VAs and proceed thereafter to embed context in the exact same way as with the FMMs. Luckily, VAs, as originally proposed [4], already respect the same simple directed graph enclosed in dashed lines in Fig. 1(a) and (b). However, priors $p(z)$, in this case, are Gaussian. Consequently, the variables $z$ cannot represent the latent class labels as we would like. It is for this reason that VAs are primarily used for dimensionality reduction–finding a compressed, informative embedding $z$ of the data $x$, but are not ideal for clustering purposes. Modeling $z$ as a discrete latent variable to get exactly the FMM graph equivalent is ill-posed because it prevents the use of the "reparameterization trick," which allows gradients to flow through random variables for parameter estimation. Although solutions have been proposed [49], it is simpler to maintain the lower dimensional embedding, here denoted $w$, and let it be governed by the latent class labels $z$, as shown in Fig. 1(c). This remains more intuitive than other approaches to implement mixture models with VAs (see [50], [51]), many of which assume a Gaussian mixture for the prior $p(w)$, instead. The mixture model in our case is thus defined with respect to the embedding $w$ rather than the data $x$, an idea introduced in [52] for semisupervised learning (M2 model). The additional contextual variables are then linked to $z$ as in the FMM case.

The profound similarity of this model's joint distribution to that of the FMMs (the only difference being the interleaving of $w$ in the message passing between $z$ and $x$) and the fact that context is embedded in the exact same place and manner makes it easy to prove that the definitions of probabilistic labels $\boldsymbol{p_i}$ are identical to the FMM case (see Appendix B). The inference model of the context-free case is $q(z, w|x) = q(z|x)q(w|x, z)$ (where $q(w|x, z)$ is the VA's encoder model) so that the context-enabled version simply consists in fusing probabilistic labels $\boldsymbol{p_i}$ with the distribution $\boldsymbol{q(z_i|x_i)}$ (softmax layer output of the corresponding network) for each sample $\boldsymbol{x_i}$, equivalently to the way shown in Table I for $p(z_i|x_i)$ in the FMM scenario (the E-step of the EM algorithm). Of note, although in Fig. 1(c), the CA case for VAs is illustrated, the WCA equivalent can be similarly derived.

Finally, the parameters of all networks defining the mixture-VA [$q(z|x)$ or $q(z|x, c)$, encoder $q(w|x, z)$, and decoder $q(x|w, z))$] are learned with gradient descent. The objective consists of the regular reconstruction loss and the regularization term for priors $p(w)$, in addition to the variational lower bound that can be written $\mathbb{E}_{q(z,w|x)}[\log p(z)/q(z|x) + \log p(w|z)/q(w|x, z) + \log p(x|z, w)]$. It must be underlined that the method is agnostic to the type of ANNs used to implement the graph.

### F. Evaluation Metrics

The results mentioned in Section IV compare five algorithms (CA, WCA, DCA, US, and S) in simulations with artificial FMM data sets, as well as in real-world problems. The estimation properties reported are precision, standard errors, and convergence rate. Estimation precision is the Euclidean distance between the estimated parameter vector $\hat{\boldsymbol{\theta}}$ and the actual one $\boldsymbol{\theta^A}$, namely, $D = \|\hat{\boldsymbol{\theta}} - \boldsymbol{\theta^A}\|$. For standard errors, we employ the aforementioned estimator $\text{SE}_i$. For brevity, we only report the average $\text{ASE} = 1/L \sum_{i=1}^{L} \text{SE}_i$ (where $L$ is the number of estimated parameters). Similarly, the MIP-based estimate $r' = 1 - r$ is used for the convergence rate. The classification performance of trained models is assessed through $N$-class accuracy $A = N_c/N$, where $N_c$ is the number of correctly classified samples out of $N$ total samples across all classes. Finally, for regression tasks, the mean square error MSE is reported.

In order to quantify the information content of side information, we employ a scaled negentropy definition on probabilistic labels $\boldsymbol{p_i}$: $\text{NE}_i = 1 + \sum_{j=1}^{M} p_{ij} \log_M p_{ij}$. This metric is conveniently bounded, $NE_i \in [0, 1]$, for any number of mixtures $M$. $\text{NE}_i = 0$ when $\boldsymbol{p_i}$ is uniform, $p_{ij} = 1/M, \forall j \in [1, M]$ ("ignorant" context, $\boldsymbol{p_i}$ does not cast a preference over any class). Conversely, $\text{NE}_i = 1$ when $p_{im} = 1, m \in [1, M]$ and $p_{ij} = 0, \forall j \neq m, j \in [1, M]$ ("perfect" context, fully revealing the class label $\boldsymbol{y_i}$). The NE level of a data set is extracted as the average across all included labels $\boldsymbol{p_i}$.

### G. Simulation Design

For our simulation studies, a label $\boldsymbol{p_i}$ for each sample $\boldsymbol{x_i}$ is constructed randomly so that its information content is $\text{NE}_i$. For all but one examined scenario, all samples in $X$ are assigned the same NE value ($\text{NE}_i = \text{NE}, \forall i$). In the "mixed" context scenario (see Appendix E in the Supplementary Material), however, each $\text{NE}_i$ is randomly drawn from a fixed interval. In all but the "wrong" context scenario, $\boldsymbol{p_i}$s are constructed to cast greater confidence to the ground-truth label $\boldsymbol{y_i}$ ("correct" context). Formally, we impose $\arg\max\{\boldsymbol{p_i}\} = \arg\max\{\boldsymbol{y_i}\}$ so that $\boldsymbol{p_i}$s always "predict" the correct $\boldsymbol{y_i}$ with increasing confidence as NE increases. This rule is only abandoned in the "wrong" context scenario, where the effects of misleading contextual information are investigated. In this scenario, $k_i = \arg\max\{\boldsymbol{p_i}\} \neq \arg\max\{\boldsymbol{y_i}\}$, $k_i \in [1, M] \forall i$ is selected randomly out of the $M - 1$ remaining possibilities for a reported percentage of the generated $\boldsymbol{p_i}$s.

The following scenarios are considered with "correct" context: A—a mixture of two univariate normal distributions, where variances are known and only the two class means are estimated, B—a mixture of two univariate normal distributions, where all existing parameters are estimated, C—a mixture of three univariate normal distributions, D—a mixture of two multivariate (2-D) normal distributions, E—a mixture of two univariate Maxwell–Boltzmann distributions, and F—a mixture of two univariate, first-order, linear regressors. These six scenarios are chosen to differ in terms of the numbers and types of mixtures employed (where the Maxwell–Boltzmann of Scenario E is not a member of the exponential family), the number of estimated parameters, the dimension of the input space, and the utility of the FMM (classification versus regression). The scenarios targeting "mixed" and "wrong" context situations (Appendix E, see the Supplementary Material) employ mixtures of two univariate normal distributions.

For each scenario, 1000 estimation problems are generated and solved for all compared algorithms. Each problem $r \in [1, 1000]$ is associated with a randomly generated data set $X_r, Y_r, P_r^{\mathrm{NE}}$ of observed data, ground-truth labels, and probabilistic labels of NE, respectively. For algorithms CA, WCA, and DCA, each problem $r$ is further solved for NE $\in [0 : 0.1 : 0.99]$ so that our evaluation encompasses the complete range of possible contextual information content. The cardinality $N$ of each data set is fixed to 100 times the number of parameters to be estimated. The ground truth $Y_r$ is constructed to have a balanced number of samples per class. The observed data $X_r$ are randomly generated from semirandomly selected "actual" distributions with parameters $\boldsymbol{\theta}_r^{\boldsymbol{A}}$ (of the respective scenario's type) and the estimation begins with semirandomly chosen initialization $\hat{\boldsymbol{\theta}}_r^{\mathbf{0}}$ (common to all algorithms). These semirandom procedures, detailed in Appendix C, ensure a balanced number of samples per class and minimal impact of separability and initialization on the extracted results. All algorithms are left to perform as many iterations $t$ as needed so that $\|\hat{\boldsymbol{\theta}}_r^{t} - \hat{\boldsymbol{\theta}}_r^{t-1}\| < 10^{-5}$. If this stopping criterion is not reached after 300 iterations for some algorithms, $\hat{\boldsymbol{\theta}}_r^{300}$ is used as its final estimate.

The classification accuracy $A$ is computed for each scenario, problem $r$ and algorithm, by generating a second "testing" data set $X_r', Y_r'$ (of equal cardinality to $X_r$) from the same "actual" FMM, which is classified using the estimated parameters of each algorithm by means of the Maximum-A posteriori rule. For the mixture-of-regressions scenario, the same evaluation methodology is applied to derive the MSE on the testing set.

## IV. Results

### A. Results With Artificial Data

Sections III-C and III-D justify theoretically the estimation benefits brought forward by the proposed approach. More specifically, in Section III-C, we show, on the one hand, how the log-likelihood objectives in this methodology, compared with the US equivalent, exhibit lower local maxima except for the one closer to the supervised estimate. This effect increases the chances of convergence to this favorable extremum by reducing the EM algorithm's sensitivity to initialization. On the other hand, Section III-D shows that the fraction of missing information of an estimation problem, as expressed by the spectral radius of the rate matrix (see Section III-B), is shown to decrease proportionally to the information content of the provided side information, bounded by US and S (maximum and zero missing information, respectively). Therefore, the estimation properties that depend on the fraction of missing information, namely, the standard errors and the convergence rate, also benefit from context awareness. Appendix D formally proves that these effects generalize to all FMMs. The theoretically anticipated effects are verified by simulations with artificially generated data, presented in this section.

The following set of simulations is meant to compare the performances of the derived algorithms for the "correct" context situation. Within each scenario, we illustrate each metric's average across all 1000 problems for algorithm $\alpha$ and some NE, $\overline{M}_{\alpha}^{NE}$, normalized to be bounded by S and US performances, as $\widetilde{M}_{\alpha}^{\mathrm{NE}} = (\overline{M}_{\alpha}^{\mathrm{NE}} - \overline{M}_{US})/(\overline{M}_{S} - \overline{M}_{US})$.

Fig. 3(a) and (b) shows that for all examined properties, CA and WCA exhibit improved performances proportionally to the strength of contextual assistance. In addition, the average performances are upper bounded by S (at NE $\to$ 1) and lower bounded by US (at NE $\to$ 0). CA is consistently outperforming WCA for the same NE, whereas both algorithms yield substantially better $D$, $A$, and/or MSE than DCA. While WCA reduces to US for NE $= 0$ (Appendix B and Section III-D), CA yields improvements over US even in this case. Scenario A is an exception occurring for fixed $\boldsymbol{\pi}$ (see Section III-D). These trends are universal although the magnitude of improvements as a function of NE is metric- and scenario-dependent, confirming the generalizability of the examples in Appendixes III-C and III-D.

Statistical testing reveals that the added value of CA and WCA over conventional unsupervised estimation US is significant for all metrics, already at very low NE (for CA, even for ignorant context). Scenarios E and F for metrics $A$ and MSE, respectively, are exceptions where improved estimation precision $D$ does not translate into significantly better classification/regression, as an intrinsic property of the respective FMMs. Nevertheless, for metrics $D$ and ASE, it is only at very high NE that CA and WCA become indistinguishably similar to S, while for $r'$, supervised learning S is significantly better even at NE $= 0.99$. The undesirable significance is notwithstanding that the average differences for all metrics tend to operate much closer to the S rather than the US "extremity" even at low NE (especially for CA). Furthermore, concerning $A$ and MSE, context-aware algorithms are statistically similar to S since very low NE. The superiority of CA over WCA is shown to be significant only for the first few tested NE levels for metrics $D$, ASE, and $A$/MSE, while it persists for almost the entirety of the NE spectrum for $r'$. Finally, DCA is again shown to underperform.

Last but not least, our simulations demonstrate that context awareness can substantially reduce the number of problems that could not converge with regular US training.[6] In the most characteristic example (due to the larger number of mixtures) of Scenario C, where 83.7% of problems did not converge with US, this percentage is reduced by CA to 3.8% at NE $= 0$ and 1.0% for NE $\in [0.1, 0.99]$. WCA also alleviates this problem (less aggressively), by gradually reducing the nonconvergence percentage to 67% at NE $= 0.1$, 39.4% at NE $= 0.5$ and, eventually, 1.0% at NE $= 0.99$. It is thus shown that context awareness is able to avoid irregularities, already at low NE levels. In the interest of space, results on more "realistic" situations where contextual assistance can be, to some extent, "wrong" and of "mixed" NE are offered in Appendix E (see the Supplementary Material).

[6]An EM algorithm fails to converge to a local maximum when the spectral radius of $J$ exceeds unity [46].

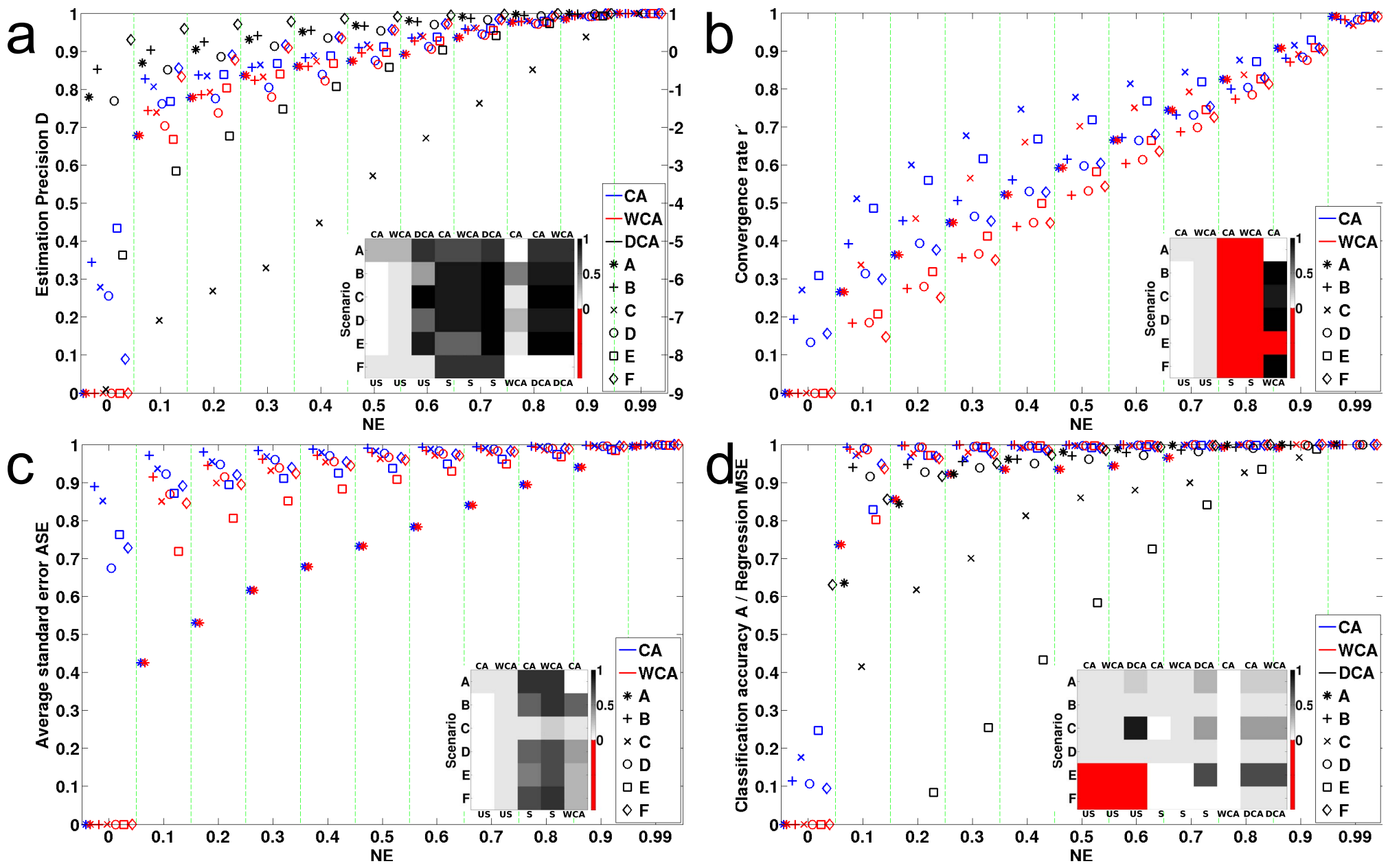

Fig. 3. Normalized, average (across 1000 problems of each of six FMM estimation scenarios A–F). (a) Estimation precision $D$ (right $y$-axis only refers to DCA performances). (b) Convergence rate $r'$. (c) Average (across parameters) standard error ASE. (d) Classification accuracy $A$/regression mean square error MSE. The performances of each context-aware algorithm (CA, WCA, and DCA) for each scenario A–F are color- and shape-coded as shown in the legends. Performances for different NE levels arranged along the horizontal axis. For each metric and scenario, the embedded colormaps illustrate the lowest (when US is involved) or highest (for all other combinations) NE level (color-coded as shown in the adjacent colorbars) above which the algorithms on top and bottom of each column significantly differ at the 95% confidence interval (one-sided, paired Wilcoxon rank-sum tests with Bonferroni correction for multiple comparisons). Red denotes no significant difference for any NE level.

TABLE II

CONCENTRATED CLASSIFICATION ACCURACY RESULTS (%) FOR FMM UNSUPERVISED CLASSIFICATION APPLICATION SCENARIOS. SCENARIO A: BCI. SCENARIO B: CANCER DETECTION. SCENARIO C: FINGERSPELLING

| | *US* | *CA* | *WCA* | *S* |
|---|---|---|---|---|
| A | $52.4 \pm 8.1$ | $64.7 \pm 6.5$ | - | $66.2 \pm 7.2$ |
| B | $79.6 \pm 8.5$ | $88.1 \pm 3.4$ | $89.8 \pm 3.1$ | $91.8 \pm 1.9$ |
| C | $61.0 \pm 9.4$ | $72.1 \pm 7.9$ | - | $90.1 \pm 2.9$ |

## *B. Results on Real-World Scenarios*

*1) Gaussian Mixture Models:* The versatility of CA and WCA parameter estimation is shown in three different unsupervised classification scenarios using the Gaussian mixture models (GMMs): 1) a brain–computer interface (BCI) speller [53], where the BCI classifier translating brain signals into text-entry actions is learned on-the-fly with CA context derived from a language model and the speller's structure [54]; 2) unsupervised CA and WCA training of classifiers diagnosing malignancies in mammograms, where known risk factors and manual assessments represent the side information at hand; and 3) unsupervised recalibration of fingerspelling classifiers recognizing sign language from video frames, where, again, a language model provides contextual assistance. Detailed methods and results on all these applications are elaborated in Appendix F (see the Supplementary Material), while Table II summarizes the average improvements obtained with the proposed algorithms against US, and how their performance compares to supervised learning S. Overall, the findings of the simulations with artificial data are shown to replicate with realistic data, the main conclusion being that the proposed context-aware learning significantly outperforms regular unsupervised learning in terms of classification accuracy and is competitive to the supervised case in spite of discarding the need for ground-truth labels.

*2) VAs and Comparison With a Neural-Symbolic Method:* The applicability of the idea put forward in this article with generative deep neural nets is shown by employing the VA-based mixture model explained in Section III-E to perform unsupervised classification of the well-known MNIST benchmark database of handwritten digits (55 000 training and 10 000 testing labeled samples for ten classes—digits 0–9). Since our goal is to show the merits of context-aware learning and not to optimize the network architecture, we vectorize the $28 \times 28$ pixel images (to 784-D input vectors) and use regular, nonconvolutional layers. Specifically, the encoder model consists of two hidden layers with 512 units each and an additional third layer with 128 units (64 + 64) to output the means and variances of the latent space $w$. The decoder model consists of three layers with 512, 512, and 784 units, whereas the model $q(z|x)$ contains three layers with 512, 512, and 10 units. RelU activation is used everywhere apart from the output layers of the decoder (logit), the $q(z|x)$ network (logit+softmax), and the encoder (no activation for means and softplus for variances, to imitate a covariance matrix).

Contextual information derives by considering a scenario of (numerical) student ID recognition as part of an automatic exam paper processing system. Due to the inherent structure of such identification systems (e.g., serial historical numbering and continuity within a university department), the place of a digit within the student ID reveals a lot of information about its potential identity, while the statistics of this kind of side information can be very easily acquired through the university's database (importantly, with view to such use, registrations can be manipulated to yield even richer information content). Specifically, we assume a realistic case of five-digit-long IDs, where the first digit is always 1, the second is 7/8/9 with probabilities 0.05/0.80/0.15, respectively, and the remaining three places yield distributions that slightly diverge from the uniform one, each favoring some of the remaining digits. For our simulations, we generate randomly such student IDs and assign MNIST samples and probabilistic labels $\boldsymbol{p_i}$ accordingly to each digit.

We compare this approach with the algorithm proposed in [42], which we consider to be the most relevant work in the ANN literature for learning with side information. This method distills additional information in the form of logic rules (hence its affinity to neural-symbolic integration). Its basic premise is an iterative KD algorithm, which augments the objective function of a regular supervised ANN with an extra term encouraging the network's softmax output to fit, in addition to the ground-truth labels $\boldsymbol{y}$, also a "soft" prediction $\boldsymbol{s}$. The latter results from the fusion of the network's output with evidence from the logic rules [42, eq. (4)], what the authors call the "teacher network" output. Of note, since rules are ultimately used to constrain $p(z|x)$, the authors admit that like we have also denoted for our own algorithms (Appendix A), this method also falls under the class of PR.

For a fair comparison based on the same model and type/level of contextual assistance, we employ the same ANN $q(z|x)$ that is part of our VA-based mixture formulation and set the soft prediction $\boldsymbol{s}$ to be the Bayesian fusion of the same probabilistic labels $\boldsymbol{p_i}$, with the network's output $\boldsymbol{q(z|x)}$ (i.e., the two distributions are multiplied and the result is normalized). Since this method explicitly balances the supervised and the rule-based objectives with a regularizer $\pi \in [0, 1]$ [see (2)], we compare with: 1) Sd: $\pi = 0$, i.e., regular supervised training of the discriminative model; 2) KD KD0.1: $\pi = 0.1$; 3) KD0.5; 4) KD0.9; 5) KD1.0 ($pi = 1.0$, and thus, ground-truth labels are ignored and the network is trained only on side information); and 6) KDexp: $\pi_t = 1 - 0.95^t$, where $t$ is the epoch. These are compared with VA-US (the softmax output $q(z|x)$ of our model is left untouched by context) and VA-CA and VA-S (class labels are fed as contextual assistance). A limited number of labeled data (only 10 per class) are used to "seed" VA-CA so as to enforce the $i$th cluster to represent the corresponding class (otherwise, it is unknown in what order the elements of probabilistic labels $\boldsymbol{p_i}$ should be presented to the algorithm), while the same data are used to associate the found clusters to the digit classes for VA-US (the class with most such samples in a cluster "wins" it).

We run ten repetitions of the learning problem (100 epochs) with each algorithm (exceptionally, 30 repetitions for VA-CA) and record the finally convergent classification accuracy of each repetition. We obtain the following results, VA-US: $71.0 \pm 9.0$, VA-CA: $82.2.0 \pm 15.4$, VA-S: $98.1 \pm 0.07$, Sd/KD0.0: $98.1 \pm 0.09$, KDexp: $49.5 \pm 0.1$, KD0.1: $98.2 \pm 0.06$, KD0.5: $97.9 \pm 0.06$, KD0.9: $86.3 \pm 0.2$, and KD1.0: $49.3 \pm 0.05$. First, it is clear that our VA-based context-aware algorithm performs equivalently to the simple FMM-based one; it significantly ($p = 0.03$ with a nonparametric, two-sided ranksum test) outperforms the corresponding VA-US algorithm and approaches the performance of VA-S. This implies that the basic idea proposed here is applicable to complex, high-dimensional problems, in which simple generative models may not be expressive enough to solve, so that deep learning is required. The fact that VA-CA is still inferior to VA-S by a large margin is due to the fact that certain repetitions still get stack to local minima (the best repetitions converge to 96%). Safely assuming that this algorithm has a similar effect on the likelihood landscape as in the FMM case (see Section III-C), stronger context should be able to alleviate this issue. However, we believe that a more expressive ANN (i.e., with more hidden layers and/or convolutional layers) may also be able to cope with this. Another important observation is that when trained in a supervised manner, the extra parameterization of the VA-based model is not detrimental to classification accuracy since this model performs similar to Sd. Hence, generative models can be equally effective for classification tasks to their discriminative counterparts while being able to do in parallel much more than that. However, it should be highlighted that this is probably only true when enough data are available, as in this exemplary case.

On the contrary, the same side information exploited through the knowledge distillation method given in [42] does not seem to improve substantially over the regular supervised case (the improvement of KD0.1 over Sd is marginal and may also be random due to the small number of repetitions performed here). The hyperparameterizations where substantial importance is given to the side-information-based objective (KDexp, KD0.9, and KD1.0) perform poorly, while those where the influence of context is reduced simply converge to the supervised performances (i.e., context seems to have no impact). This was somewhat anticipated since the weak information provided on the data labels through context seems redundant when strong information of the actual labels is available. Of note, the results in [42] do not question this conclusion because the benefits demonstrated thereby are also slight and likely attributed to the fact that manual labeling in the cases examined there (i.e., sentiment analysis) is rather ambiguous and noisy (thus, ground-truth information is also weak, allowing a margin of improvement through logic rule constraints). This should be considered in addition to the fact that the approach proposed here has anyway the advantage over knowledge distillation of being unsupervised. A final important observation regards the compromised performance of KD1.0. Our FMM-based results suggested that the DCA method underperforms (unless very strong contextual assistance is available), which led us to confirm that data evidence carries a lot of information that should not be ignored. Extending this conclusion, since KD1.0 performs much worse than VA-US, it seems that even if data evidence is employed, this will not be enough if the underlying model is discriminative and does not model the data distribution.

## V. Discussion

This work has studied unsupervised MLE algorithms devoid of any need for data labels but able to exploit side information in the form of probabilistic context embedded into a generative model and with known statistics. A comparative analysis and in-depth study of these algorithms' properties for FMMs are offered from both a theoretical and a practical standpoint. An implementation with VAs extending the same idea to deep learning is also provided.

We argue that the literature of learning from side information has largely ignored the merits of such fundamental techniques in favor of alternatives mostly stemming from the theory and practice of undirected graphs, which can often be less practical. As an example, we find that in most application scenarios and for most prospective users, it should be possible and much more intuitive to express a given type of side information as a random variable with known statistics linking it to the missing class labels, rather than through a GEC or a "feature expectation." In other words, the admittedly greater generality and flexibility of PR can be most often traded off (Appendix A). Simplicity of the resulting derivations and formulations can only be viewed as additional advantages. We thus consider bringing this methodology in the spotlight to be the main contribution of this article.

Another contribution entails the identification of basic principles giving rise to improved EM-MLE by context awareness. First, we have shown that a context-assisted log-likelihood objective is favorably distorted in comparison to the regular one so that sensitivity to initialization diminishes and the chances of convergence closer to the supervised MLE increase. The second principle regards the partial elimination of missing label information through context as a result of the applicability of the MIP. Demonstrating this makes our work the first one to justify the benefits of side information in learning from an information-theoretic viewpoint. Through these principles, we have established experimentally and, wherever possible, also formally, two important points. First, any positive effects on the estimation properties are proportional to the information content of implicitly extracted instancewise probabilistic labels. Second, the proposed algorithms perform between the boundaries defined by the unsupervised and supervised equivalents of a given problem.

We have showcased that estimation benefits are still evident and significant in problems with variable, weak, or even, to a certain extent, "wrong" contextual assistance, situations likely to arise in practical applications. Furthermore, the algorithms' limitations and comparative advantages have been outlined. In this regard, we have demonstrated that completely disregarding the evidence from observed samples in favor of context, such as with the DCA algorithm, or attempting to exploit it when the underlying model is not generative, yields inferior estimation properties. A general superiority of CA over WCA as a result of removing missing information related to the mixing coefficients has also been demonstrated. Furthermore, the application of all algorithms in tough, real-world problems, and extensibility to deep learning scenarios, showcases the broad applicability of context-aware learning as proposed here.

As argued in Section III, the main limitation of the algorithms proposed here is their nonuniversal applicability. However, this is not specific to the proposed framework, but, rather, a limitation shared among all methods exploiting side information. Indeed, it is not guaranteed that for any application, exploitable context as shown here exists or that the cost of automatically retrieving contextual assistance will be lower than that of explicitly labeling data. However, as the real-world examples of this article illustrate, rich context should be easily and cheaply acquired in a broad application spectrum. Another limitation regards the fact that the prerequisite knowledge of distributions $p(z|c)$ or $p(c|z)$ implies that the latent class labels are at least defined, i.e., the number and type of mixtures/classes $M$ are known. Consequently, the proposed methodology regards unsupervised classification and not general clustering problems.

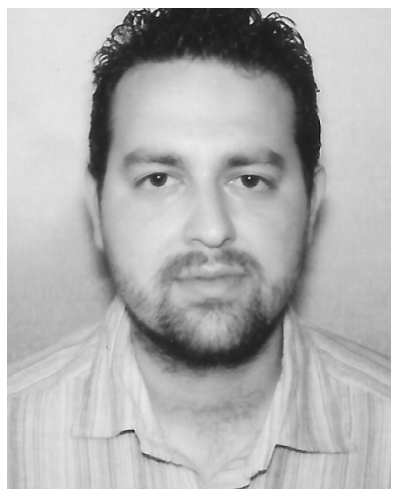

**Serafeim Perdikis** (Member, IEEE) received the M.E. degree in electrical and computer engineering from the Aristotle University of Thessaloniki (AUTH), Thessaloniki, Greece, in 2007, and the Ph.D. degree in brain-computer interaction (BCI) from the École Polytechnique Fédérale de Lausanne (EPFL), Geneva, Switzerland, in 2014.

In 2015, he joined the Wyss Center for Bio- and Neuro-Engineering, Geneva. In November 2017, he was a Research Engineer with Mindmaze SA, Lausanne, Switzerland, where he worked on neurorehabilitation. Since February 2019, he has been a Lecturer with the Computer Science and Electronic Engineering Department, University of Essex, Colchester, U.K., and a Researcher with the Brain-Computer Interfaces and Neural Engineering Group.

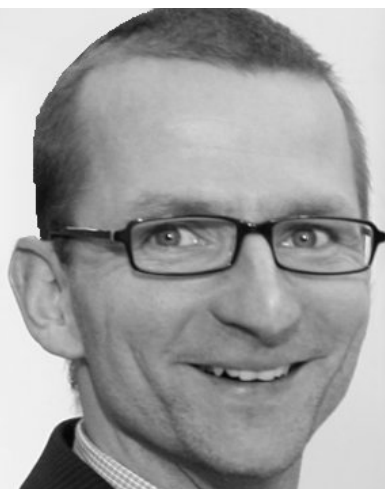

**Robert Leeb** (Member, IEEE) received the M.Sc. degree in electrical and biomedical engineering and the Ph.D. degree in computer science from the Graz University of Technology, Graz, Austria, in 2000 and 2008, respectively.

He was a Scientist and a Chief-Engineer at the Center for Neuroprosthetics, École Polytechnique Fédérale de Lausanne (EPFL), Sion, Switzerland, where he worked in applying brain-computer interface (BCI) technologies in stroke rehabilitation, and on the transfer of BCI technology toward the rehabilitation clinic SUVA-CRR. Furthermore, he was a Senior Post-Doctoral Researcher at the Chair in Non-Invasive Brain-Machine Interface, EPFL. He is leading the Bio-Signal Department, MindMaze SA, Lausanne, Switzerland. His group is responsible for algorithm development, machine learning, and bio-signal data analysis for clinical and basic research studies on rehabilitation, gaming, and motorsports. His research interests include BCI, brain stimulation, neuro-rehabilitation, bio-signal processing, hybrid and multimodal human–computer interaction, and virtual reality. He has published more than 50 articles in international, peer-reviewed journals and more than 100 conference papers. He has authored several book chapters. He holds a CAS in Advanced Studies in Management of Biotech, Medtech & Pharma Ventures from EPFL, in 2016.

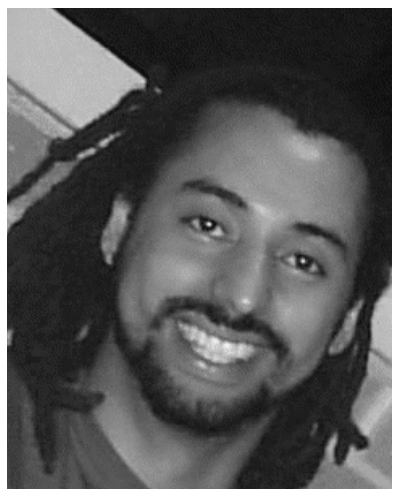

**Ricardo Chavarriaga** (Member, IEEE) has more than 12 years of experience in research on human–machine interaction, brain–machine interfaces, and artificial intelligence. His work is focused on responsible development of technologies that promote beneficial, humane interaction between human and intelligent machines. He is currently the Head of the Swiss office of the Confederation of Laboratories for AI Research in Europe (CLAIRE), Zurich, Switzerland, a Senior Researcher with the Zurich University of Applied Sciences (ZHAW), Winterthur, Switzerland, and an Associate Fellow with the Geneva Center for Security Policy (GCSP), Geneva, Switzerland.

Dr. Chavarriaga is also the Chair of the IEEE Standards Association Industry Connection Group on Neurotechnologies for Brain-Machine Interfacing.

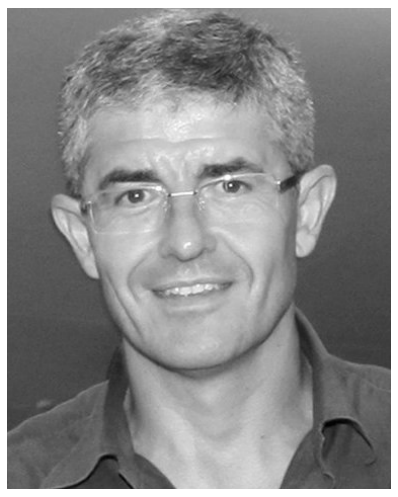

**José del R. Millán** (Fellow, IEEE) received the Ph.D. degree in computer science from the Technical University of Catalonia, Barcelona, Spain, in 1992.

He was a Research Scientist with the Joint Research Centre of the European Commission, Ispra, Italy, and a Senior Researcher with the Idiap Research Institute, Martigny, Switzerland. From 2009 to 2019, he was with the Defitech Foundation Chair in Brain-Machine Interface, École Polytechnique Fédérale de Lausanne (EPFL), Geneva, Switzerland, where he helped establish the Center for Neuroprosthetics. He has been a Professor with the Department of Electrical and Computer Engineering and the Department of Neurology, The University of Texas at Austin, Austin, TX, USA, since September 2019, where he holds the Carol Cockrell Curran Endowed Chair. He has made several seminal contributions to the field of brain–computer interfaces (BCI). In addition to his work on the fundamentals of BMI and design of neuroprosthetics, he is prioritizing the translation of BMI to end users suffering from motor and cognitive disabilities. In parallel, he is designing BMI technology to offer new interaction modalities for able-bodied people.

Dr. Millán has received several recognitions for these pioneering achievements, notably the IEEE-SMC Nobert Wiener Award in 2011 and an elected fellow of the International Academy of Medical and Biological Engineering in 2020.